\pdfoutput=1

\documentclass[11pt]{article}

\usepackage[]{EMNLP2022}

\usepackage{times}
\usepackage{latexsym}

\usepackage[T1]{fontenc}

\usepackage[utf8]{inputenc}

\usepackage{microtype}

\usepackage{graphicx}
\usepackage{amsfonts}
\usepackage{booktabs}
\usepackage{multirow}
\usepackage{amssymb}
\usepackage{amsmath}
\usepackage[utf8]{inputenc}
\usepackage{cleveref}
\crefname{section}{§}{§§}
\usepackage{subfigure}
\usepackage{array}
\usepackage{color}

\usepackage{hyperref}

\title{TSGP: Two-Stage Generative Prompting for Unsupervised Commonsense Question Answering}

\author{Yueqing Sun, Yu Zhang\thanks{\ \ Corresponding author.}, Le Qi, Qi Shi\\
Research Center for Social Computing and Information Retrieval\\
Harbin Institute of Technology, Harbin, China\\
        \texttt{\{yqsun,qshi,lqi,zhangyu\}@ir.hit.edu.cn}\\}
        
\begin{document}
\maketitle
\begin{abstract}
Unsupervised commonsense question answering requires mining effective commonsense knowledge without the rely on the labeled task data.
Previous methods typically retrieved from traditional knowledge bases or used pre-trained language models (PrLMs) to generate fixed types of knowledge, which have poor generalization ability.
In this paper, we aim to address the above limitation by leveraging the implicit knowledge stored in PrLMs and propose a two-stage prompt-based unsupervised commonsense question answering framework (TSGP). 
Specifically, we first use knowledge generation prompts to generate the knowledge required for questions with unlimited types and possible candidate answers independent of specified choices. Then, we further utilize answer generation prompts to generate possible candidate answers independent of specified choices.
Experimental results and analysis on three different commonsense reasoning tasks, CommonsenseQA, OpenBookQA, and SocialIQA, demonstrate that TSGP significantly improves the reasoning ability of language models in unsupervised settings\footnote{Our code is available at: \href{https://github.com/Yueqing-Sun/TSGP}{https://github.com/Yueqing-Sun/TSGP}}.
\end{abstract}

\section{Introduction}
\begin{figure}[t]
\centering
\includegraphics[width=1\columnwidth]{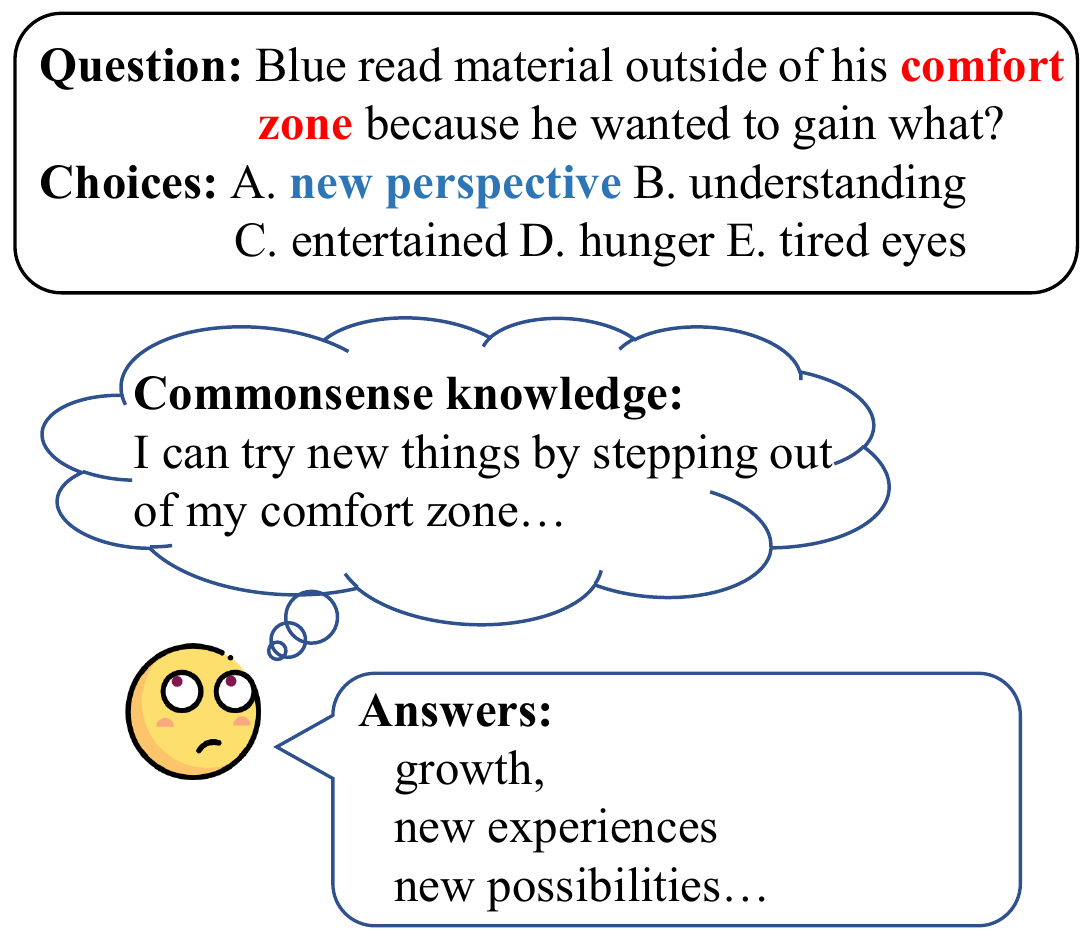} 
\caption{Humans can immediately use commonsense knowledge of life to give multiple possible answers to a commonsense question. Unsupervised CSQA models should simulate human behaviour and thinking.}
\label{fig_example}
\end{figure}

Commonsense question answering (CSQA) requires systems to acquire different types of commonsense knowledge, consisting of widely known facts that humans use to reason and respond to everyday situations, but are challenging for machines \cite{talmor-etal-2019-commonsenseqa, sap-etal-2019-social}.
As illustrated in Figure \ref{fig_example}, humans consider their own experience or education when answering questions, and their answers can vary from person to person.

Existing studies commonly focus on acquiring relevant knowledge by retrieving external knowledge bases and then fine-tuning pre-trained language models (PrLMs) in a supervised manner on task-specific data~\cite{lin-etal-2019-kagnet, feng-etal-2020-scalable, yasunaga-etal-2021-qa}. 
Recently, unsupervised commonsense question answering has attracted the attention of researchers because it does not rely on any labeled downstream task data and does not require fine-tuning of pre-trained models.

Unsupervised commonsense question answering presents two challenges: (i) eliciting the required commonsense knowledge from a pre-trained language model; and (ii) scoring the answers reasonably.
Correspondingly, previous works can be divided into two categories. One is to use preset templates to allow PrLMs to generate specific types of knowledge or to generate the meaning of questions \cite{shwartz-etal-2020-unsupervised, bosselut2021dynamic}. However, the type of knowledge generated by this method is limited, making it difficult to transfer to new task domains.
Another class of methods design scoring functions to rank answer choices, such as using cross-entropy or mutual information to score each answer choice directly or calculating the semantic similarity between choices and generated pseudo-answers \cite{niu-etal-2021-semantic}. However, they are limited to modeling using the implicit parameters of PrLMs, without explicitly capturing question-related knowledge from PrLMs and using them to guide inference. 

To alleviate the above limitations, in this paper, we propose a general prompt-based unsupervised commonsense question answering framework called Two-Stage Generative Prompting (TSGP). 
We design knowledge and answer generation prompts that can prompt the language model to flexibly generate commonsense knowledge required for questions and their multiple corresponding candidate answers.
Under the assistance of the prompts, our TSGP can generate unrestricted types of knowledge, making implicit intermediate reasoning steps explicit to bridge the gap between questions and choices. Furthermore, our TSGP can generate many diverse candidate answers independently of the fixed choices.


Specifically, we design simple prompts for knowledge generation and answer generation. First, we use knowledge prompts to let the language model generate some question-conditioned knowledge statements. The generated knowledge statements may contain noise since we do not fine-tune the language model. So we only select the most relevant piece of knowledge by calculating the mutual information between the generated knowledge and the question. Then, we leverage the language model again to generate pseudo-answers conditioned on questions and knowledge based on the answer prompts. Finally, we compute the generated pseudo-answers with semantic scores for each choice, and vote for the most semantically relevant answer. 
Experimental results on three commonsense QA benchmarks containing CommonsenseQA, OpenBookQA and SocialIQA demonstrate the effectiveness of our proposed TSGP framework.

Our contributions in this paper are summarized as follows:
\begin{itemize}
\item We propose a general two-stage generative prompting framework (TSGP) to fully exploit the knowledge implicit in PrLMs for unsupervised commonsense question answering.

\item We design knowledge and answer generation prompts in TSGP to make implicit intermediate reasoning steps explicit and generate possible candidate answers independent of specified choices.

\item We conduct experiments on three question answering datasets focusing on different types of commonsense, and find that TSGP significantly improves the reasoning ability of language models in unsupervised settings.
\end{itemize}

\begin{figure*}[t]
\centering
\includegraphics[width=1\textwidth]{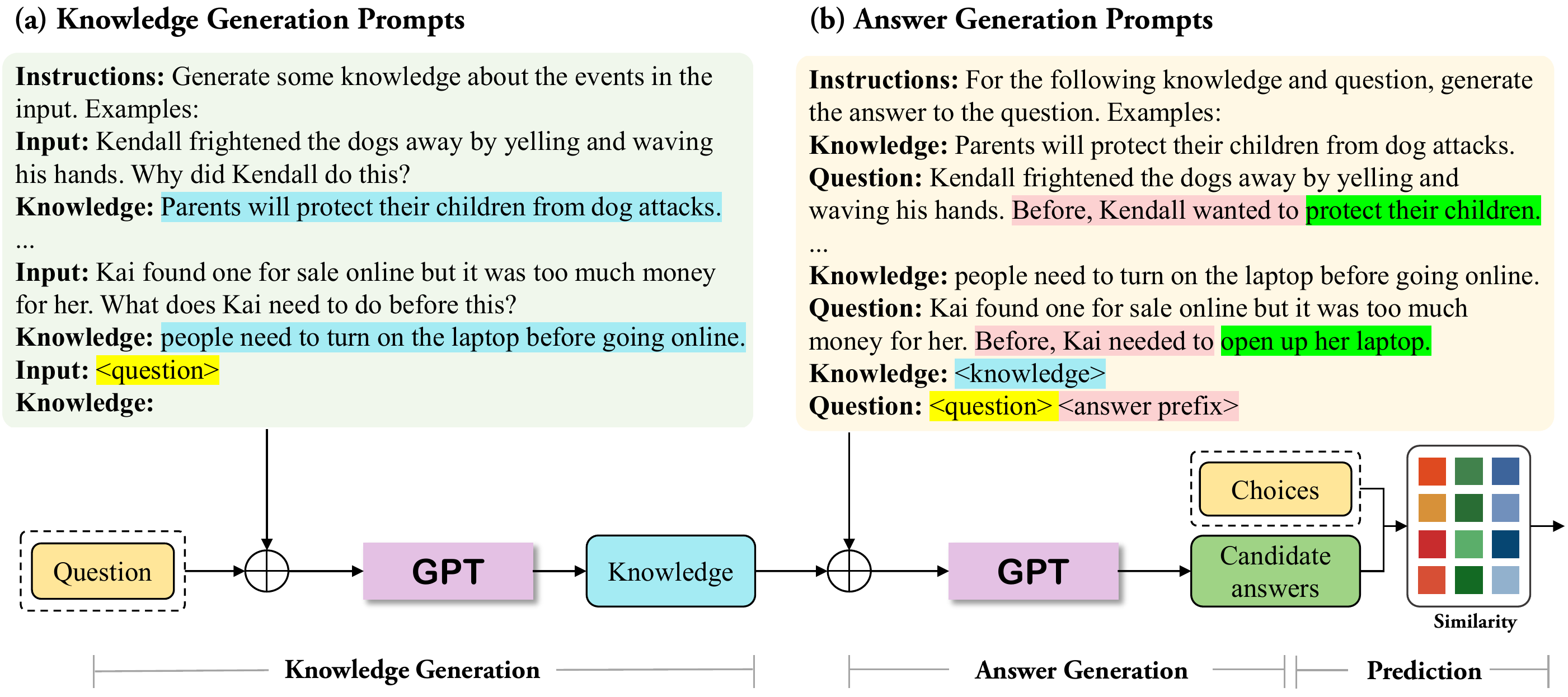} 
\caption{Overview of the proposed method \textbf{T}wo-\textbf{S}tage \textbf{G}enerative \textbf{P}rompting (TSGP), which contains three parts. 1) Knowledge Generation, using GPT to generate knowledge statements about concepts or events in questions based on the knowledge generation prompts. 2) Answer Generation, leveraging GPT again to generate pseudo-answers conditioned on questions and knowledge based on the answer generation prompts. 3) Answer Prediction, computing the semantic score between the generated pseudo-answer and each choice and voting for the final answer.}
\label{fig_model}
\end{figure*}

\section{Method}
We focus on the unsupervised multiple-choice commonsense question answering task, which is formalized as follows: given a question $q$ and a set of choices $A$, the model should choose the correct option $\hat{a}$:
\begin{equation}
\hat{a} = \underset{a\in A}{\text{argmax}} P(a|q).
\end{equation}
As shown in Figure \ref{fig_model}, we aim to fully exploit the knowledge encoded in PrLMs and propose the TSGP framework, that mainly contain 3 parts: 1) Knowledge Generation, that using GPT to generate question-conditioned knowledge statements; 2) Answer Generation, that use a language model to generate pseudo-answers conditioned by the question and the generated knowledge; and 3) Answer Prediction, that compute the semantic scores of the generated pseudo-answers with each option, and vote for the most semantically relevant option. In this subsection, we will describe these in detail.


\subsection{Knowledge Generation}
We first use knowledge generation prompts to let GPT generate knowledge (a series of coherent short sentences) related to the concepts or events in the question. Its essence is to fill knowledge gaps and clarify implicit intermediate reasoning steps to arrive at answers to reasoning questions. Instead of generating answers directly, we generate short sentences that imply the relationship between the concepts in the question and the answer, similar to a thought process or solution. 

Figure \ref{fig_model}(a) shows the design of the knowledge generation prompt for the SocialIQA dataset. The knowledge generation prompt consists of instruction, some demos fixed for each task, and a new question placeholder. Demonstrations are hand-written, and each demonstration contains a task-style question and a knowledge statement that helps answer that question. When generating knowledge for a new question q, we insert the question into a placeholder and repeatedly sample the generated continuation of that prompt to obtain a set of knowledge statements $K_q = \{k_1, k_2, ..., k_m\}$.

Since we did not train the knowledge generator to generate coherent and factually correct knowledge, it can be assumed that some of the generated knowledge did not provide helpful information for the model. So we propose to use point-wise mutual information (PMI) to quantify the correlation between question $q$ and each generated knowledge $k$.
\begin{equation}
PMI(q;k) = log\frac{p(q|k)}{p(k)} = log\frac{p(k|q)}{p(q)}
\end{equation}

Since both $q$ and $k$ are sentences, we use a PrLM to estimate the probabilities. Conditional probabilities are calculated by treating conditional sentences as prefixes. It is important to note that while PMI can by definition be computed in two equivalent ways, language model estimates do not guarantee:
\begin{equation}
\frac{PrLM(q|k)}{PrLM(k)} = \frac{PrLM(k|q)}{PrLM(q)}
\end{equation}
Therefore, we use question $q$ as the condition to calculate the mutual information between the knowledge and the question.
\begin{equation}
PMI(k;q) \overset{\mathrm{def}}{=}  log\frac{PrLM(k|q)}{PrLM(q)}
\end{equation}
It measures the dependency between questions and knowledge. A positive score means that the question is positively related to this piece of knowledge. A score of zero means that the question and this piece of knowledge are independent. A negative score means that the question and this piece of knowledge may be contradictory. We only select the piece of knowledge with the largest mutual information.
\begin{equation}
\hat{k} = \underset{k\in K_q}{\text{argmax}} PMI(k;q)
\end{equation}

\subsection{Answer Generation}
In order to take full advantage of the commonsense reasoning potential of the pre-trained language model, we did not simply use the language model to score the concatenation of knowledge and questions. Further, we design the answer generation prompts, again using the language model to generate possible fake answers. 

Our answer prompts are much the same as knowledge prompts, including an instruction, some demos fixed for each task, a new knowledge placeholder, and a new question placeholder, as shown in Figure \ref{fig_model}(b). Each demo is aligned with the knowledge generation prompt and contains the same question and knowledge statement. When generating an answer for a new question $q$, we insert the question and knowledge generated in the previous step into placeholders and repeatedly sample the generated continuation of that prompt to obtain a set of answers $S_q = \{s_1, s_2, ..., s_n\}$. Especially, to ensure the generation's quality, the question may needs to be rewritten for different tasks, and the answer prefix is added after the question.

\subsection{Answer Prediction}
After getting a series of fake answers generated by the language model, we use them to score each option. The standard practice of scoring for previous models is to concatenate each answer option $a_i$ with the question $q$, and then use the cross-entropy loss of the sentence modeled by the language model as the answer score \cite{shwartz-etal-2020-unsupervised}.
\begin{equation}
\hat{a}=\underset{a_i\in A}{\text{argmax}} \mathcal{L}_{\text{PrLM}}(q, a_i)
\end{equation}

Although the language model is successful in scoring, it will be affected by many interfering factors such as word frequency and sentence structure. These factors can interfere with the scoring function to a large extent. 

In order to mitigate the influence of these interference factors, we borrowed the practice of SEQA \cite{niu-etal-2021-semantic} to score options from the perspective of semantic similarity between pseudo-answers and options, and finally selected the option with the highest score as the answer.
\begin{equation}
P(a_i|S) = \frac{1}{n \cdot Z(T)} \sum_{j=1}^{n}{exp[\frac{cos(h_{s_j},h_{a_i})}{T}]}
\end{equation}
where $a_i$ represents the i-th option corresponding to the question, and $s_j$ represents the generated j-th pseudo-answer. $h_{a_i}$ represents the semantic representation of option $a_i$, $h_{s_j}$ represents the semantic representation of the j-th pseudo-answer, and $cos(,)$ represents the cosine similarity. $T$ is the temperature, and $Z(T) = exp()$ is the standardization term.

\section{Experiment}
In this section, we first describe our experiments on three commonsense question answering datasets, followed by further analysis and case studies.
\subsection{Datasets}
We evaluate the consistency improvement of our model using three different commonsense question answering datasets. Since the test sets are hidden, we report all results on the dev sets. Note that the labels are not visible and are only used for the final accuracy evaluation.

\textbf{CommonsenseQA} \cite{talmor-etal-2019-commonsenseqa} is a multiple-choice question answering task that requires commonsense knowledge to reason, with a total of 12,102 pieces of data (training/validation/test: 9741/1221/1140). Each piece of data contains one question and five candidate answers. These questions and candidate answers are constructed using entities in ConceptNet and aim to explore potential commonsense relationships between entities.

\textbf{OpenBookQA} \cite{mihaylov-etal-2018-suit} is a question answering dataset that mimics the open book exam for assessing human understanding of a subject. It consists of 5957 multiple-choice elementary science questions (training/validation/test: 4957/500/500). It is designed to explore the understanding of a small "book" of 1326 core scientific facts and the application of facts to new situations. Answering questions requires other extensive public knowledge than is covered in this book.

\textbf{SocialIQA} \cite{sap-etal-2019-social} is a question answering benchmark that tests commonsense intelligence on social interactions. It focuses on reasoning about people's behaviour and social impact. It contains more than 37,000 question-answer pairs used to evaluate a model's ability to reason about the social impact of everyday events and situations.

\subsection{Compared Methods}
We investigate the impact of our TSGP framework by comparing with the following methods:

\noindent\textbf{Baseline}
Our baseline model is constructed to assign each answer option $a_i$ to a combination of context and question, using only a pre-trained language model (GPT2) as a scorer without explicit knowledge injection.

\noindent\textbf{Self-talk} \cite{shwartz-etal-2020-unsupervised} acquires knowledge from PrLM through a two-stage generation. It uses a preset template to prompt PrLM to generate Information Search Questions (ISQs), the ISQs will be put back a second time, prompting PrLM to generate their answers as a clarification. Although this method is strictly unsupervised, the preset template does not Universal and must be carefully designed for different tasks.

\noindent\textbf{SEQA} \cite{niu-etal-2021-semantic} applies PrLM to generate hundreds of pseudo-answers and compares them with each option, but it does so only once with PrLM and does not take full advantage of the implicit knowledge in PrLM.

\noindent\textbf{DynaGen} \cite{bosselut2021dynamic} uses COMET \cite{bosselut-etal-2019-comet} to generate intermediate inferences, which are then used to score choices. 
\noindent\textbf{GKP} \cite{liu-etal-2022-generated}
selects some samples from the task data to prompt the knowledge required for GPT-3 \cite{brown2020language} model generation, then splices each generated knowledge and question into LM for scoring.

For Self-talk and SEQA, we rerun their code on GPT-2 at different scales using the parameter settings provided by the authors and report the results of our reruns. For DynaGen and GKP, we only report the results on the same dataset given in the paper, since we do not have access to the model source code or have permission to use GPT-3. 
\begin{table*}[t]
\centering
\resizebox{2\columnwidth}{!}{
\begin{tabular}{l|l c c c c|c}
\toprule   
\textbf{Datasets} & \textbf{Models} & GPT-2 Small & GPT-2 Medium  & GPT-2 Large & GPT-2 XL & Published\\
\hline
\multirow{5}{2.6cm}{\textbf{CommonsenseQA}} & Baseline &  29.0 &  29.1 &  32.6 &  32.3 &  --\\
 & Self-talk &  24.8 &  27.3 &  31.5 &  31.4&  32.4$^\dagger$ \\
 & SEQA &  26.1 &  30.7 &  34.6 &  34.8 & -- \\
 & GKP & -- &  -- &  -- &  -- &  47.3$^\ddagger$\\
 & TSGP (Ours) & 33.3 &  42.2 &  46.8 &   \textbf{49.1}& -- \\
\hline
\multirow{4}{2.6cm}{\textbf{OpenBookQA}} & Baseline &  16.4 &  18.0 &  20.0 &  22.8 &  --\\
 & Self-talk &  17.4 &  21.0 &  23.8 &  25.4 &  --\\
 & SEQA &  27.6 &  28.6 &  32.0 &   33.4& -- \\
 & TSGP (Ours) & 38.0 & 43.8 &  43.0 &  \textbf{44.4} & -- \\
\hline
\multirow{5}{2.7cm}{\textbf{SocialIQA}} & Baseline &  39.8 &  41.8 &  43.0 &  42.8 &  --\\

 & Self-talk &  41.2 &  43.3 &  45.3 &  46.2 &  46.2\\
 & SEQA &  44.4 & 44.6  & 46.6  &  47.5 &  47.5\\
 & DynaGen & -- &  -- &  -- &  -- &  50.1$^\diamondsuit$\\
 & TSGP (Ours) & 45.9 &  46.7 &  49.7 &  \textbf{51.5} &  --\\
\bottomrule
\end{tabular}
}
\caption{Accuracy (\%) on CommonsenseQA, OpenBookQA and SocialIQA. The best results are depicted in boldface. All results except the last column are run by ourselves. ‘$^\dagger$’ indicates that Self-talk uses GPT2-Medium to generate knowledge, and then uses GPT2-Large to predict answers. ‘$^\ddagger$’ indicates that GKP uses GPT-3 to generate knowledge, and then uses T5-11b to make answer predictions.  ‘$^\diamondsuit$’ indicates that DynaGen uses COMET to generate knowledge and inferences.}
\label{main_results_table}
\end{table*}

\begin{table}[t]
\centering
\resizebox{1.0\columnwidth}{!}{
\begin{tabular}{lcc}
\toprule   \textbf{Methods} & \textbf{CommonsenseQA} & \textbf{OpenBookQA} \\
\hline
GPT-2 XL  & 32.3 & 22.8\\
\quad+ Knowledge Generation	& 45.5 & 30.0\\
\quad+ Answer Generation	& 43.9 & 42.0\\
\quad+ Both (Ours) & \textbf{49.1} & \textbf{44.4}\\
\bottomrule
\end{tabular}
}
\caption{Ablation study on model components.}
\label{components}
\end{table}

\subsection{Implementation Details}
The examples included in two-stage prompts are randomly sampled from the training set of the dataset. In order to ensure the generality of the framework, we do not carefully design and select the prompts. See the appendix \ref{sec:appendix} for all prompts.
We use GPT-2 as the baseline model. In order to obtain reliable and reproducible results, we conduct four different scale experiments on GPT-2: GPT-2 Small (117M), GPT-2 Medium (345M), GPT-2 Large (762M) and GPT-2 XL (1.5B). For knowledge generation, we generate M = 20 knowledge statements per question, kernel sampling p = 0.5, and discard duplicates and empty strings. Generation terminates when more than 64 tokens are exceeded, or a `.' token is hit. For question generation, we use GPT-2 with the same size as in knowledge generation to generate candidate answers by kernel sampling p = 0.9, and the sample size N of candidate answers is set to 500. Following the setting of SEQA, we use SRoBERTa-large \cite{reimers-gurevych-2019-sentence}, which is further fine-tuned on the NLI dataset, to calculate the semantic similarity between the generated candidate answers and the individual options. We set the parameter T = 0.1 on the validation set of all three datasets.

\begin{table*}[t]
\centering
\resizebox{2\columnwidth}{!}{
\begin{tabular}{l|p{8.5cm}ll}
\toprule  
\textbf{\textbf{Model}}        & {\textbf{Question} / \textit{Generated Knowledge} / Generated Answer}                       & \multicolumn{1}{l}{\textbf{Predicted}} & \textbf{Score}       \\
\hline
Baseline              & \textbf{Which large land mass is home to the most monkeys?}                                           & amazon basin (×)                        & 0.35                  \\ \hline
\multirow{2}{*}{TSGP} & \textit{The world's largest monkey colony lives on Madagascar.}                                       & \multirow{2}{*}{african continent ($\checkmark$)}  & \multirow{2}{*}{0.51} \\ \cline{2-2}
                      & “Madagascar, Africa”, “Africa”, “Madagascar, Democratic Republic of the Congo” …             &                                         &                       \\ \hline
Baseline              & \textbf{What green area is a marmot likely to be found in?}                                           & north america (×)                       & 0.21                  \\ \hline
\multirow{2}{*}{TSGP} & \textit{Marmots live near water sources so they need plenty of vegetation around their habitat.}      & \multirow{2}{*}{countryside ($\checkmark$)}        & \multirow{2}{*}{0.34} \\ \cline{2-2}
                      & “rural area”, “country side”, “village” …                                                    &                                         &                       \\ \hline
Baseline              & \textbf{Where do all animals live?}                                                                   & zoos (×)                                & 0.43                  \\ \hline
\multirow{2}{*}{TSGP} & \textit{Animals exist on every continent except Antarctica (where they're frozen).}                   & \multirow{2}{*}{surface of earth ($\checkmark$)}   & \multirow{2}{*}{0.46} \\ \cline{2-2}
                      & “the oceans and land”, “all over the earth”, “wherever they can find the food and shelter” … &                                         &                       \\ 
\bottomrule
\end{tabular}
}
\caption{Some examples where our model corrects the baseline (GPT2-XL) predictions. The first line of each part is the original question and the prediction result of the baseline; the second line is the knowledge statement generated by TSGP, and the third line is the pseudo-answer generated by TSGP. We show the correct answer with a check mark ($\checkmark$) and the wrong answer with a wrong mark (×).}
\label{case_table}
\end{table*}

\subsection{Main Results}
Table \ref{main_results_table} shows the results for the three benchmarks. Our TSGP model achieves the best performance among the unsupervised models on all datasets. We outperform baseline by 16.8\% on the CommonsenseQA dataset and 5.2\% higher than the previous best model SEQA; on OpenBookQA On SocialIQA, we outperform baseline by 21.6\% and 10.2\% higher than the previous best model SEQA; on SocialIQA, we outperform baseline by 8.7\% and 4.0\% higher than the previous best model SEQA. 
Furthermore, our model steadily brings positive improvements over the baseline models on GPT at different scales. 

Compared to our model, Self-talk fails to maintain effectiveness, and its accuracy is sometimes even slightly lower than the baseline, especially on the CommonsenseQA dataset, suggesting that the knowledge generated by Self-talk may be noisy and mislead model evaluation.
With the help of SRoBERTa-NLI large, SEQA can achieve better results on these three datasets, but it does not fully exploit the implicit knowledge encoded in GPT and is less effective than our model.
We also observed a common phenomenon in our TSGP, baseline, and comparison models: when GPT-2 was improved from Large (750M) to Xlarge (1500M), the model performance did not improve as much as expected. We speculate that there may be limitations to using only model parameters to improve the performance of language models as sentence evaluators or generators.

\subsection{Analysis}
In this subsection, we perform more analysis to show the impact on the performance of two-stage generative prompting and the sample size of generated answers. We then perform the human evaluation of the quality of the generated knowledge and answers. Finally, we present some case studies.
\paragraph{Ablation Studies}
To identify the source of the performance increase, we perform an ablation study on TSGP. 

As shown in Table \ref{components}, the baseline model GPT2-XL performs poorly and cannot answer common sense questions well. Our models outperform the baseline by 8\% to 20\% on all tasks. It can be seen that both knowledge prompts and answer prompts are significantly improved compared to the baseline model, and our framework achieves the best results after integrating the two. Furthermore, through knowledge generation, our method can effectively derive commonsense knowledge implicit in GPT parameters, helping to improve performance on commonsense question answering tasks, where knowledge is an essential factor.

\paragraph{Sample Size for Generating Answers}
Then, we conduct an comparison experiment to investigate the effect of the generated answer sample size N on accuracy.

\begin{figure}[t]
\centering
\includegraphics[width=1\columnwidth]{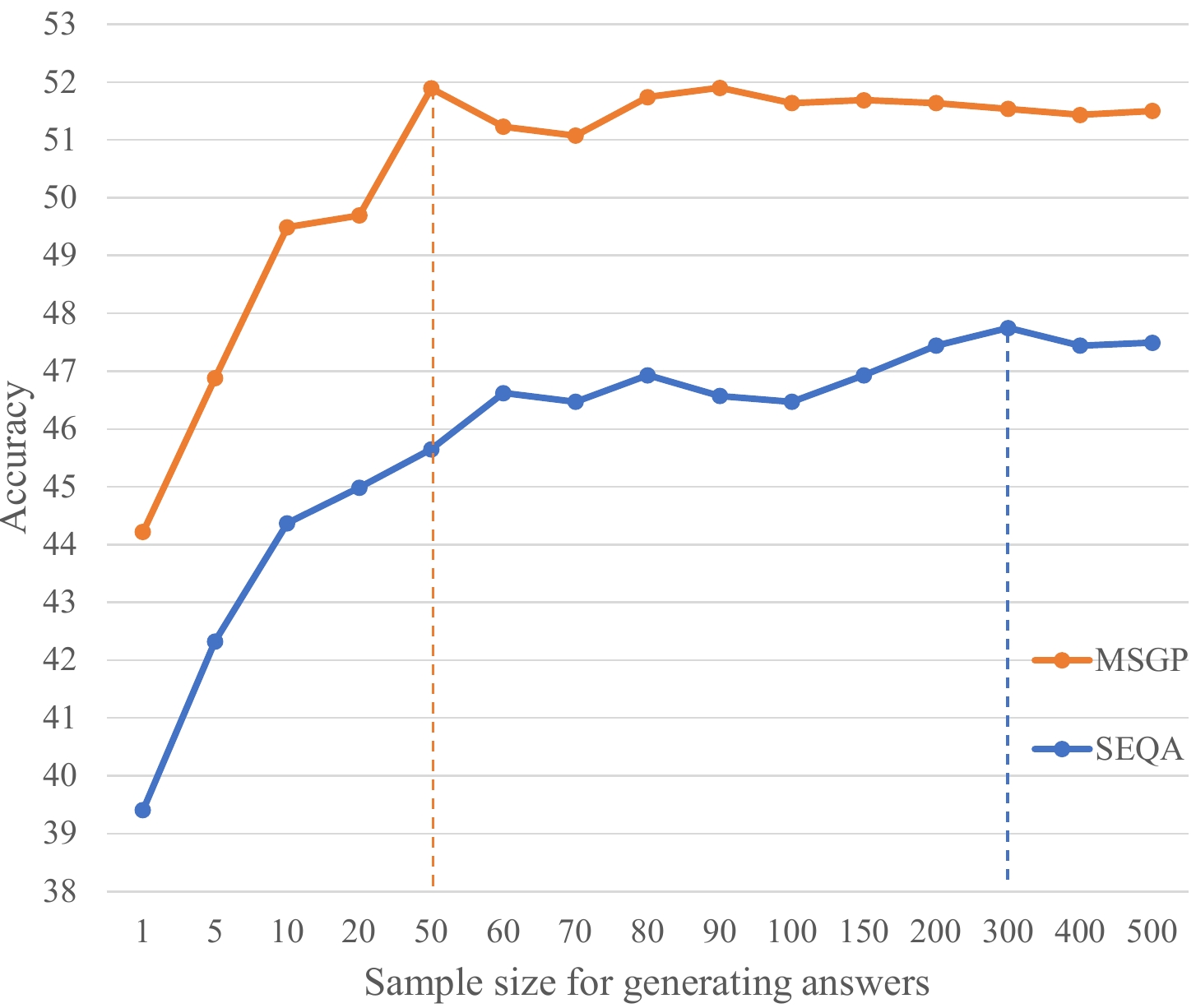} 
\caption{The effect of the number of generated answers on the accuracy. Our TSGP has the highest accuracy with an answer sample size of 50, while SEQA needs to generate more answers to achieve high accuracy.}
\label{sample_answers}
\end{figure}

As shown in Figure \ref{sample_answers}, we compared our TSGP and SEQA models. Both TSGP and SEQA use GPT2-XL to generate answers, but TSGP also uses GPT2-XL to generate explicit knowledge. As expected, accuracy increases with sample size. Specifically, for TSGP, the performance improves rapidly when K < 50, and then the growth slows down or almost no growth when K > 50. Ultimately, TSGP achieves stable and relatively high performance. It is clear that TSGP only needs to generate fewer answers to achieve better accuracy than SEQA, proving that previous models do not fully exploit the knowledge implicit in language model parameters. Explicitly generating the knowledge encoded in the model parameters can, in turn, guide the language model to reason in the direction of the answer.

\paragraph{Human evaluation}
We conduct human evaluations to investigate the interpretability of the quality of generated knowledge and generated answers and their impact on task performance. 

We evaluate the quality of the generated knowledge in three dimensions: 1) Grammatical: whether it is grammatical; 2) Relevant: whether it is relevant to the topic or concept mentioned in the question; 3) Usefulness: whether it is helpful to directly or indirectly way to answer questions. We only use relevance and usefulness to evaluate the generated answers. Since the generated candidate answers are generally words or phrases and rarely complete sentences, we do not evaluate the answers for grammar. 

\begin{figure}[t]
\centering
\includegraphics[width=0.9\columnwidth]{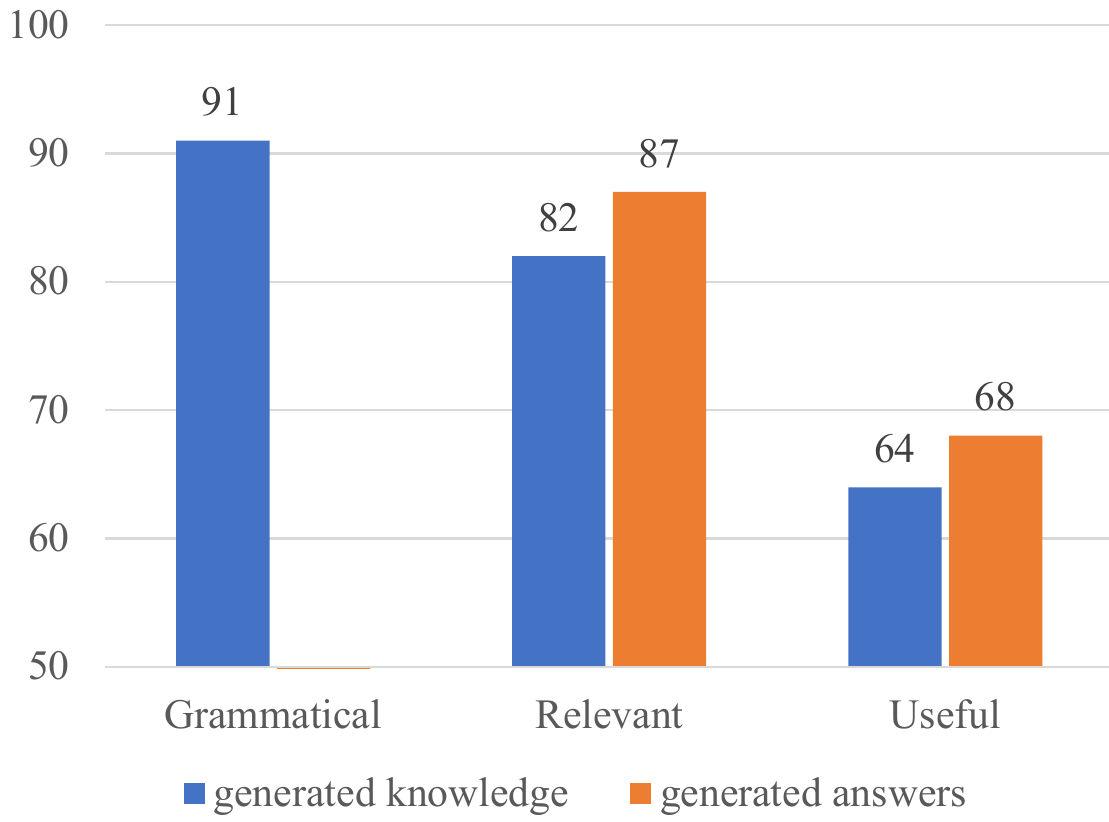} 
\caption{Human evaluation of generated knowledge and generated answers.}
\label{human}
\end{figure}

We sample 100 generated knowledge and 100 generated answers from the CommonsenseQA dataset, and the evaluation results are shown in Figure \ref{human}. 91\% of the generated knowledge conforms to the basic grammar, 82\% is related to the entities in the question, but only 64\% contributes to the answer. 87\% of the generated answers were question-related, slightly higher than knowledge, suggesting that knowledge can guide the model to reason toward the answer. However, only 68\% of the answers generated are useful, indicating that there is still much room for improvement in guiding the model to answer common sense questions in an unsupervised setting. At the same time, we noticed that the usefulness of the answers from the human evaluation is lower than the accuracy of the model prediction. The reason may be that many commonsense questions have divergent answers, and reasonable answers may not be within the given five options.

\paragraph{Case Study}
Finlly, we perform a case study to more intuitively demonstrate the changes that the knowledge generated by our framework and the generated answers bring to model predictions. 

We select some examples from the CommonsenseQA dataset, as shown in Table \ref{case_table}. 
These examples come from the intersection of the baseline model's predictions being wrong and our framework's predictions being correct, i.e. our framework correcting the baseline model's wrong predictions. In all examples, the baseline model rates incorrect answers higher than correct answers, while under our framework, correct answers are rated much higher. Part of the reason is that the generated knowledge can transform commonsense reasoning into explicit reasoning, such as paraphrasing, induction, and analogy. Moreover, part of the reason is that we use answer hints to generate candidates that are semantically similar to the answer, making the model no longer limited by Character-level matching, while focusing more on semantic information.

\section{Related Work}
\subsection{Extracting knowledge from pre-trained language models}

Recent studies have shown language models trained on corpora large enough to implicitly encode many different types of knowledge in their parameters so that they can act as knowledge bases \cite{petroni-etal-2019-language, tenney2019you}. Compared with traditional KBs, this method has a significant advantage in requiring no human supervision. However, there is also a problem that KBs can be easily accessed by querying specific entity nodes, but LMs are more challenging to access specific knowledge pieces.

\citet{brown2020language} find that LMs can be effective few-shot and zero-shot learners, and the knowledge learned before training can be acquired through fine-tuning or prompting.
Fine-tuning LMs on downstream tasks has proven to be an effective way to tune and acquire specific knowledge for evaluation \cite{dong2019unified, da2021analyzing}. However, the ever-increasing size of LMs makes them expensive to fine-tune and store in practice.
On the other hand, Prompting is an efficient way to directly capture the needed knowledge from LMs without any additional fine-tuning \cite{davison-etal-2019-commonsense, jiang2020can, liu2021pre, roberts-etal-2020-much}. This paradigm provides models with a familiar query format, resulting in better responses \cite{alkhamissi2022review}.

\subsection{Unsupervised commonsense question answering}

Previous work has explored pre-trained language models for unsupervised commonsense question answering tasks.
DynaGen \cite{bosselut2021dynamic} uses COMET \cite{bosselut-etal-2019-comet} to generate intermediate inferences, which are then used to score choices. However, its use of COMET as a generator limits its applicability to the domain of social commonsense.
Self-Talk \cite{shwartz-etal-2020-unsupervised} queries GPT \cite{radford2019language} through task-specific templates to extract knowledge. Therefore, Self-Talk can be applied to many domains but requires careful handcrafting to generalize to new tasks.
SEQA \cite{niu-etal-2021-semantic} utilizes generative PrLM to generate hundreds of pseudo-answers, then computes the semantic similarity between candidates and these generated pseudo-answers and votes for the answer. However, since there are no task examples to prompt, SEQA may need to generate hundreds of pseudo-answers for reasoning to achieve good results.
\citet{liu-etal-2022-generated} proposed GKP, which selects some samples from the task data to prompt the knowledge required for GPT-3 \cite{brown2020language} model generation, then splices each generated knowledge and question into LM for scoring. Unlike previous methods of acquiring knowledge using task-specific templates or fine-tuning knowledge generators, GKP only requires a few human-written presentation in task styles, making it very flexible, easily transferable, and engineering efficiency.

We extend GKP, and the difference is that
1) we design the prompt for answer generation so that GPT can generate multiple candidate answers, reducing the competition of surface form among options; 2) We use a similarity-based method for answer reasoning. The influence of statistical deviations such as word frequency and length of options is reduced; 3) Our framework can achieve good results by using GPT-2, much smaller than GPT-3, and improving parameter efficiency.

\section{Conclusion}
In this paper, we propose a two-stage prompt-based unsupervised commonsense question answering framework that makes implicit intermediate reasoning steps explicit and generates possible candidate answers independent of specified choices. 
Specifically, we design knowledge and answer generation prompts that can prompt the language model to flexibly generate commonsense knowledge required for a question and its multiple corresponding candidate answers. 
Experiments show our method's effectiveness across multiple datasets, achieving state-of-the-art results on three commonsense reasoning tasks in an unsupervised setting. In addition, it does not require fine-tuning, does not depend on specific pre-trained models and tasks, and is flexible in transfer. We hope that our framework will facilitate future research in this field.

\section*{Limitations}
We explore an unsupervised commonsense question answering framework that leverages knowledge encoded in pre-trained language models. It uses a pre-trained language model to generate knowledge and multiple possible answers for commonsense questions based on multi-stage prompts. As for the limitations, the first is that the design of the prompts may need to be optimized. The examples contained in the two-stage prompts of MSGP are randomly sampled from the train set of the dataset. In order to ensure the generalization of the framework, we have not carefully designed and selected the prompts. The second is the limitation of computing resources. MSGP would theoretically perform better on larger models, such as GPT-3.

\section*{Acknowledgements}
We would like to thank the anonymous reviewers for their helpful comments. This work was supported by the National Natural Science Foundation of China (No.61976068) and "Hundreds, Millions" Engineering Science and Technology Major Special Project of Heilongjiang Province (No.2020ZX14A02).

\section*{Ethics Statement}
This paper proposes a general framework for unsupervised commonsense question answering. We worked within the purview of acceptable privacy practices and strictly followed the data usage policy. In all the experiments, we use public datasets and consist of their intended use. We neither introduce any social/ethical bias to the model nor amplify any bias in the data, so we do not foresee any direct social consequences or ethical issues.

\bibliography{anthology,custom}
\bibliographystyle{acl_natbib}

\appendix

\section{Appendix}
\label{sec:appendix}
Tables \ref{CSQA_prompt_knowledge}, \ref{OBQA_prompt_knowledge}, and \ref{SIQA_prompt_knowledge} are the Knowledge Generation Prompts corresponding to the CommonsenseQA, OpenBookQA, and SocialIQA datasets. Tables \ref{CSQA_prompt_answer}, \ref{OBQA_prompt_answer}, and \ref{SIQA_prompt_answer} are the Answer Generation Prompts corresponding to the CommonsenseQA, OpenBookQA, and SocialIQA datasets.

\begin{table*}[p]
\centering
\resizebox{2\columnwidth}{!}{
\begin{tabular}{l|p{12cm}}
\toprule   
\textbf{CommonsenseQA} & Instructions: Generate some knowledge about the concepts in the input. Examples:\\ \cline{1-1}
\multirow{5}{2.6cm}{\textbf{Knowledge Generation Prompts}} 
&    Input: Google Maps and other highway and street GPS services have replaced what?\\
&    Knowledge: Electronic maps are the modern version of paper atlas.\\
&    Input: The fox walked from the city into the forest, what was it looking for?\\
&    Knowledge: Natural habitats are usually away from cities.\\
&    Input: You can sharefiles with someone if you have a connection to a what?\\
&    Knowledge: Files can be shared over the Internet.\\
&    Input: Too many people want exotic snakes. The demand is driving what to carry them?\\
&    Knowledge: Some people raise snakes as pets.\\
&    Input: The body guard was good at his duties, he made the person who hired him what?\\
&    Knowledge: The job of body guards is to ensure the safety and security of the employer.\\
& Input: \\
& Knowledge: \\
\bottomrule
\end{tabular}
}
\caption{Knowledge Generation Prompts of CommonsenseQA.}
\label{CSQA_prompt_knowledge}
\end{table*}

\begin{table*}[p]
\centering
\resizebox{2\columnwidth}{!}{
\begin{tabular}{l|p{12cm}}
\toprule   
\textbf{OpenBookQA} & Instructions: Generate some knowledge about the concepts in the input. Examples:\\ \cline{1-1}
\multirow{5}{2.6cm}{\textbf{Knowledge Generation Prompts}} 
& Input: As you look deeper into a marbel you can see?\\
& Knowledge: As the size of an object appears larger, that object will be observed better.\\
& Input: In the wilderness, light pollution is?\\
& Knowledge: As distance to a city decreases, the amount of light pollution will increase.\\
& Input: Earth rotating causes?\\
& Knowledge: A planet rotating causes cycles of day and night on that planet.\\
& Input: Renewable resources?\\
& Knowledge: Renewable resources can be used over again.\\
& Input: The removal of trees may cause damage to ecosystems such as?\\
& Knowledge: Cutting down trees has a negative impact on an organisms living in an ecosystem.\\
& Input: \\
& Knowledge: \\
\bottomrule
\end{tabular}
}
\caption{Knowledge Generation Prompts of OpenBookQA.}
\label{OBQA_prompt_knowledge}
\end{table*}

\begin{table*}[p]
\centering
\resizebox{2\columnwidth}{!}{
\begin{tabular}{l|p{12cm}}
\toprule   
\textbf{SocialIQA} &      Instructions: Generate some knowledge about the events in the input. Examples:\\ \cline{1-1}
\multirow{5}{2.6cm}{\textbf{Knowledge Generation Prompts}}
&    Input: Kendall frightened the dogs away by yelling and waving his hands. Why did Kendall do this?\\
&    Knowledge: Parents will protect their children from dog attacks.\\
&    Input: Cameron decided to have a barbecue and gathered her friends together. How would others feel as a result?\\
&    Knowledge: Gathering with friends for a barbecue is a great pleasure.\\
&    Input: Kendall ran back and thanked Lee for helping her find the dog. How would you describe Kendall?\\
&    Knowledge: We are grateful that others have helped us.\\
&    Input: Jan needed to give out jobs for an upcoming project at work. What will Others want to do next?\\
&    Knowledge: After being assigned a new project people will start working.\\
&    Input: Kai found one for sale online but it was too much money for her. What does Kai need to do before this?\\
&    Knowledge: People need to turn on the laptop before going online.\\
& Input: \\
& Knowledge: \\
\bottomrule
\end{tabular}
}
\caption{Knowledge Generation Prompts of SocialIQA.}
\label{SIQA_prompt_knowledge}
\end{table*}

\begin{table*}[p]
\centering
\resizebox{2\columnwidth}{!}{
\begin{tabular}{l|p{12cm}}
\toprule   
\textbf{CommonsenseQA} & Instructions: For each question below, guided by the knowledge, choose the answer from the answer bank corresponding to the question that best answers the question.\\ \cline{1-1}
\multirow{5}{2.6cm}{\textbf{Answer Generation Prompts}}
& Knowledge: Electronic maps are the modern version of paper atlas.\\
& Question: Google Maps and other highway and street GPS services have replaced what? Atlas.\\
& Knowledge: Natural habitats are usually away from cities. \\
& Question: The fox walked from the city into the forest, what was it looking for? Natural habitat.\\
& Knowledge: Files can be shared over the Internet. \\
& Question: You can share files with someone if you have a connection to a what? Computer network. \\
& Knowledge: Some people raise snakes as pets. \\
& Question: Too many people want exotic snakes. The demand is driving what to carry them? Pet shops. \\
& Knowledge: The job of body guards is to ensure the safety and security of the employer. \\
& Question: The body guard was good at his duties, he made the person who hired him what? Feel safe. \\
& Knowledge: \\
& Question: \\
\bottomrule
\end{tabular}
}
\caption{Answer Generation Prompts of CommonsenseQA.}
\label{CSQA_prompt_answer}
\end{table*}

\begin{table*}[p]
\centering
\resizebox{2\columnwidth}{!}{
\begin{tabular}{l|p{12cm}}
\toprule   
\textbf{OpenBookQA} &  Instructions: For the following knowledge and question, generate the answer to the question.\\ \cline{1-1}
\multirow{5}{2.6cm}{\textbf{Answer Generation Prompts}} 
& Knowledge: As the size of an object appears larger , that object will be observed better.\\
& Question: As you look deeper into a marbel you can see minut defects.\\
& Knowledge: A tape measure is used to measure length.\\
& Question: With which could you tell the exact size of an object? A plastic tape with graduated markings.\\
& Knowledge: A planet rotating causes cycles of day and night on that planet.\\
& Question: Earth rotating causes the cycling of AM and PM.\\
& Knowledge: Mammals give birth to live young.\\
& Question: Which animal gives birth to live young? Giraffe.\\
& Knowledge: Cutting down trees has a negative impact on an organisms living in an ecosystem.\\
& Question: The removal of trees may cause damage to ecosystems such as jungles.\\
& Knowledge: \\
& Question: \\
\bottomrule
\end{tabular}
}
\caption{Answer Generation Prompts of OpenBookQA.}
\label{OBQA_prompt_answer}
\end{table*}

\begin{table*}[p]
\centering
\resizebox{2\columnwidth}{!}{
\begin{tabular}{l|p{12cm}}
\toprule   
\textbf{SocialIQA} &  Instructions: For the following knowledge and question, generate the answer to the question.\\ \cline{1-1}
\multirow{5}{2.6cm}{\textbf{Answer Generation Prompts}} 
& Knowledge: Parents will protect their children from dog attacks.\\
& Question: Kendall frightened the dogs away by yelling and waving his hands. Before, Kendall wanted to protect their children.\\
& Knowledge: Gathering with friends for a barbecue is a great pleasure.\\
& Question: Cameron decided to have a barbecue and gathered her friends together. As a result, Others felt like attending.\\
& Knowledge: We are grateful that others have helped us.\\
& Question: Kendall ran back and thanked Lee for helping her find the dog. Kendall is seen as grateful.\\
& Knowledge: After being assigned a new project people will start working.\\
& Question: Jan needed to give out jobs for an upcoming project at work. As a result, Others wanted to get to work.\\
& Knowledge: Business restaurants are usually located in business sector.\\
& Question: Kai found one for sale online but it was too much money for her. Before, Kai needed to open up her laptop.\\
& Knowledge: \\
& Question: \\
\bottomrule
\end{tabular}
}
\caption{Answer Generation Prompts of SocialIQA.}
\label{SIQA_prompt_answer}
\end{table*}

\end{document}